\documentclass{Interspeech}

\interspeechcameraready 
\usepackage{todonotes}
\usepackage{tcolorbox}
\usepackage{caption}
\usepackage{footmisc}
\usepackage{fix-cm}

\title{Granary: Speech Recognition and Translation Dataset in 25 European Languages}

\author[affiliation={1}]{Nithin Rao}{Koluguri$^{\dagger*}$}
\author[affiliation={2}]{Monica}{Sekoyan$^{*}$}
\author[affiliation={2}]{George}{Zelenfroynd$^{*}$}
\author[affiliation={3}]{Sasha}{Meister$^{*}$}
\author[affiliation={1}]{Shuoyang}{Ding$^{*}$}
\author[affiliation={2}]{Sofia}{Kostandian}
\author[affiliation={1}]{He}{Huang}
\author[affiliation={2}]{Nikolay}{Karpov$^{\#}$}
\author[affiliation={1}]{Jagadeesh}{Balam}
\author[affiliation={1}]{Vitaly}{Lavrukhin}
\author[affiliation={4}]{Yifan}{Peng}
\author[affiliation={5}]{Sara}{Papi}
\author[affiliation={5}]{Marco}{Gaido}
\author[affiliation={5}]{Alessio}{Brutti}
\author[affiliation={1}]{Boris}{Ginsburg}

\affiliation{}{NVIDIA}{USA}
\affiliation{}{NVIDIA}{Armenia}
\affiliation{}{NVIDIA}{Germany}
\affiliation{}{Carnegie Mellon University}{USA}
\affiliation{}{Fondazione Bruno Kessler (FBK)}{Italy}

\email{nkoluguri@nvidia.com$^{\dagger}$, nkarpov@nvidia.com$^{\#}$}


\usepackage{comment}
\usepackage{graphicx}
\usepackage{multicol}
\usepackage{multirow}
\usepackage{float}
\usepackage{amsmath}
\usepackage{xcolor}
\usepackage{footmisc}
\definecolor{myblue}{RGB}{115, 147, 195}
\definecolor{mygreen}{RGB}{217, 234, 211}
\usepackage{cite}
\usepackage{url}
\usepackage{boxedminipage}
\usepackage{amssymb}
\usepackage{svg}
\usepackage{inconsolata}

\newif\ifcondition
\conditionfalse

\begin{document}

\maketitle
\footnotetext[1]{\textit{The starred(*) authors contributed equally to this work.}}
\footnotetext[2]{\textit{Corresponding authors: Nithin Rao Koluguri$^{\dagger}$, Nikolay Karpov$^{\#}$}}

\begin{abstract}
Multi-task and multilingual approaches benefit large models, yet speech processing for low-resource languages remains underexplored due to data scarcity. To address this, we present Granary, a large-scale collection of speech datasets for recognition and translation across 25 European languages. This is the first open-source effort at this scale for both transcription and translation. We enhance data quality using a pseudo-labeling pipeline with segmentation, two-pass inference, hallucination filtering, and punctuation restoration. We further generate translation pairs from pseudo-labeled transcriptions using EuroLLM, followed by a data filtration pipeline. Designed for efficiency, our pipeline processes vast amount of data within hours.  We assess models trained on processed data by comparing their performance on previously curated datasets for both high- and low-resource languages. Our findings show that these models achieve similar performance using approx. 50\% less data. Dataset will be made available at \url{https://hf.co/datasets/nvidia/Granary}.

\end{abstract}

\section{Introduction}
Advancements in speech transcription and translation technologies have been propelled by the increasing availability of large-scale datasets. These systems, which underpin applications such as automatic speech recognition (ASR) and automatic speech translation (AST), require extensive and diverse data to achieve high accuracy, robustness, and scalability. The necessity for such data arises from the complexity of human speech, which encompasses a vast range of linguistic, acoustic, and contextual variations.

Despite the growing demand, high-quality human-annotated speech data remains scarce due to the high cost and extensive effort required for curation. Unlike textual data, the availability of human-annotated speech data is significantly constrained, posing challenges for the continued development of speech foundation models. With the rise of large language models (LLMs), substantial computational resources have been allocated to training such systems, and projections suggest that human-generated text annotations may soon become depleted \cite{villalobos2022will}. A similar trend is expected for human-labeled speech data.

However, a vast amount of unlabeled speech data exists online, offering an opportunity to enhance speech models through pseudo-labeling techniques. This is particularly critical for low-resource languages, where manually annotated speech data is even scarcer. By leveraging pseudo-labeled data, ASR and AST systems can be significantly improved for underrepresented languages, mitigating linguistic biases and fostering more inclusive speech technologies.

While pseudo-labeled data is increasingly utilized in speech model training \cite{barrault2023seamless,puvvada2024less}, much of this data remains proprietary. Open-sourcing such datasets would promote transparency, reproducibility, and accessibility in speech research, facilitating broader collaboration between academia and industry. This is particularly important for low-resource languages, where public access to high-quality training data could accelerate the development of more accurate speech models.

Efforts to open-source speech data remain limited. Notable examples include YODAS \cite{li2024yodasyoutubeorienteddatasetaudio} and YouTube-Commons (YTC)
\cite{pleias2025youtube}, which provide large-scale datasets with labels derived from YouTube captions, albeit without guarantees regarding quality or source reliability. More recently, MOSEL \cite{mosel} has released pseudo-generated labels for European languages, covering datasets such as VoxPopuli \cite{wang-etal-2021-voxpopuli} and LibriLight \cite{kahn2020libri}. Other community efforts have highlighted corpus creation pipelines, but these remain restricted to human-generated data and cover only a limited number of languages \cite{chen2021gigaspeechevolvingmultidomainasr}.

Aside from ASR transcripts, open-source projects tackling translation tasks—particularly in speech applications—are exceptionally sparse. Pseudo-label generation for such tasks typically relies on training text-based neural machine translation models to produce automatic speech translation (AST) pairs. However, recent advancements in LLMs have significantly improved their reliability for these tasks.
Motivated by similar effort in text translation \cite{finkelstein-etal-2024-introducing}, we explore the use of open-source LLMs for generating pseudo-labeled translation pairs for speech translation, which is the first to the best of our knowledge.
Our approach builds on prior ASR and AST\cite{barrault2023seamless,puvvada2024less} pseudo-labeling efforts by improving the efficiency of the labeling pipeline, ensuring open-source accessibility, expanding language coverage, and generalizing across diverse corpora. 

To summarize, the main contributions of this work are as follows:  
\begin{itemize}  
    \item Open-source large-scale speech processing pipeline.\footnote{\href{https://github.com/NVIDIA/NeMo-speech-data-processor/tree/main/dataset_configs/multilingual/granary}{\url{https://github.com/NVIDIA/NeMo-speech-data-processor/tree/main/dataset_configs/multilingual/granary}}}
    
    \item Efficient method for generating translation pairs from ASR transcripts across 25 languages.
    \item 643k hours of high-quality pseudo-labeled data for 25 languages.
    \item Evaluation of the quality of pseudo-labeled data against the MOSEL pipeline for both high- and low-resource languages.
\end{itemize}

\begin{figure*}[!ht]
  \centering
  \caption{Granary pseudo-labeling pipeline. The pipeline consists of two parts: ASR and AST. The ASR pipeline includes segmentation, two-pass ASR model inference, language ID verification, text filtering, and Punctuation and Capitalization (PnC) restoration. The AST pipeline involves AST pair generation using the EuroLLM model, followed by Quality Estimation filtering.}
  \includegraphics[width=0.88\linewidth]{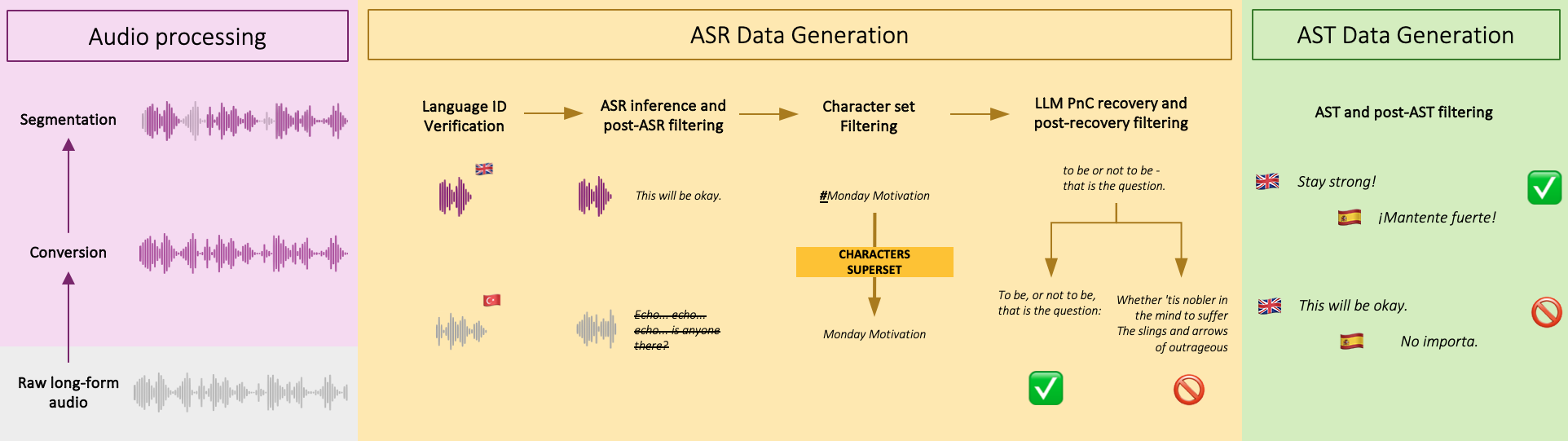}
  \label{fig:speech_production}
\end{figure*}

\section{Data}

\label{sec:data}
In this section, we describe the datasets used for pseudo-labeling.
\sloppy

This work focuses on 25 languages (23 EU languages, Ukrainian, and Russian). The EU languages include: Bulgarian (bg), Czech (cs), Danish (da), German (de), Greek (el), English (en), Spanish (es), Estonian (et), Finnish (fi), French (fr), Croatian (hr), Hungarian (hu), Italian (it), Lithuanian (lt), Latvian (lv), Maltese (mt), Dutch (nl), Polish (pl), Portuguese (pt), Romanian (ro), Slovak (sk), Slovenian (sl), and Swedish (sv).

We consider three major open-source Creative Commons speech corpora: YODAS \cite{li2024yodasyoutubeorienteddatasetaudio}, YouTube-Commons (YTC) \cite{pleias2025youtube}, and MOSEL \cite{mosel}. Each presents challenges in annotation quality, noise, and language distribution. Table \ref{tab:corpora_hrs} lists unfiltered hours and language coverage.

YODAS\cite{li2024yodasyoutubeorienteddatasetaudio}, a large-scale multilingual dataset with over 500k hours in 100+ languages, derives annotations from YouTube subtitles, which are often unreliable. Even manually created captions lack guaranteed human verification. Language ID inaccuracies lead to significant data loss (e.g., only 20\% retention for Bulgarian, Ukrainian), necessitating robust filtering. Additionally, the dataset contains noise, requiring extensive preprocessing.

YTC\cite{pleias2025youtube}, similar to YODAS, sources transcriptions from YouTube captions, inheriting reliability issues. It is heavily skewed toward English (70\% of data), limiting multilingual applications. Due to download constraints, only a subset is currently processed, with the remainder planned for future work.

MOSEL\cite{mosel} comprises of VoxPopuli \cite{wang-etal-2021-voxpopuli} and LibriLight \cite{kahn2020libri}, pseudo-labeled using Whisper-large-v3 \cite{radford2023robust}. However, transcription errors, particularly truncated segments, compromise completeness and require correction mechanisms.

\begin{table}[b]
    \centering
    \caption{Language coverage and total number of hours for each Granary corpora before and after filtration pipeline.}
    \resizebox{\linewidth}{!}{%
    \begin{tabular}{c|c|c|c|c}
       \toprule
       \textbf{Corpora} & \textbf{Languages} & \textbf{Unfiltered Hours} & \textbf{Filtered Hours} & \textbf{Retention Rate [\%]} \\
       \toprule
        \midrule YODAS & 23 & 363,549.3 & 192,172.16 & 52.86 \\
        YTC & 24 & 255,333.72 & 122,474.77 &  47.9 \\
        MOSEL & 23 & 440,712.51 & 328,590.64 & 74.56\\
        \midrule
        Total & 25 (Unique) & 1,059,595.53 & 643,237.57 & 60.7 \\
         \bottomrule
    \end{tabular}%
    }
    \label{tab:corpora_hrs}
\end{table}

\section{Granary Pipeline}

Figure 1 presents the generic pipeline, divided into two main parts: data preparation for ASR and separately for AST. To provide a better understanding of each step, its content, and the underlying experiments, we will discuss them in detail in the following subsections.

\subsection{ASR Data Pipeline}

Building on prior research \cite{mosel}, we identified Whisper-large-v3 \cite{radford2023robust} as a strong candidate for pseudo-labeling due to its robust performance, multilingual capabilities, and open license. However, its direct application requires careful adjustments and filtering due to several challenges. Whisper exhibits reduced accuracy in low-resource languages and is prone to hallucinations, particularly in its turbo variant. It struggles with noise and non-speech segments, necessitating a robust voice activity detection (VAD) system. Additionally, language identification errors, fixed 30-second segment requirements, and lack of case control in output text further complicate its use. 
Addressing these limitations is crucial for effectively leveraging Whisper for pseudo-labeling leading us to design Granary pipeline which we will outline in this section.

All files were converted to FLAC or WAV formats at a sample rate of 16 kHz and mono-channel to ensure consistency. Additionally, we set a maximum duration of 40 seconds for the final audio files\cite{koluguri2024longer}.

\subsubsection{Long-form Audio Segmentation}
\label{subsec:long-form-segmentation}

The availability of ground truth transcriptions in the YouTube data necessitated the use of an alignment algorithm to segment the audio and assign the corresponding transcriptions to each segment. We experimented with multiple alignment methods, including VAD, NeMo Forced Alignment (NFA)\cite{rastorgueva2023nemo}, Time-Duration-Transducer (TDT)\cite{koluguri2024longer} decoder and Whisper timestamps.

Using ASR models (Parakeet\cite{rekesh2023fast} \& Whisper) for timestamp generation, we compared ground truth and intermediate transcripts, finding that pseudo-labels consistently improved segmentation results. Thus, we adopted them for data processing in the Granary corpus (evaluation results omitted for space constraints). When evaluating segmentation methods, we found no significant differences in final model performance. Therefore, we chose Whisper model timestamps for the YODAS and YTC sets and Parakeet\footnote{\url{https://hf.co/nvidia/parakeet-tdt_ctc-110m}} model timestamps for the LibriLight dataset. Further analysis showed that Whisper's segment-level timestamps were poor at word alignment but effective for speech/non-speech detection. This led us to run a second-pass inference with Whisper to generate transcripts for newly segmented audio. In contrast, TDT Decoder timestamps provided high-quality segment-level alignment, eliminating the need for a second inference pass for approximately 60k hours of En data.

\begin{figure}[htbp]
    \centering
    \caption{Example of MOSEL Transcription Truncation Issue and the same sentence transcription with Granary Pipeline.}
    \begin{tcolorbox}[
        colframe=black,
        colback=white,
        arc=2mm,
        width=\linewidth,
        boxrule=0.2mm,
    ]
    \textbf{MOSEL Transcription:}  

    a Council recommendation on the strengthening cooperation against vaccine-preventable diseases. Thank you very much for your attention. \textcolor{red}{{[MISSING PART]}}

    \vspace{0.5em}
    \textbf{Granary Transcription:}  

    a Council recommendation on the strengthening cooperation against vaccine-preventable diseases. Thank you very much for your attention. \textcolor[rgb]{0,0.5,0}{{And thank you very much, Madam. Colleagues, our debate is closed.}}
    \end{tcolorbox}
    \label{fig:mosel_transcription}
\end{figure}

\subsubsection{Two-Pass Inference}

Building on the previous subsection, where we highlighted the need for generating a high volume of pseudo-labels—even for ground truth samples—we initiated the Whisper-large-V3 pipeline using FasterWhisper\cite{faster-whisper} with a beam size of 5 and a chunk batch of 16. Following MOSEL's best practices \cite{mosel}, we performed two-pass inference: first for language ID prediction, then for transcription, using the predicted language ID as metadata to improve data quality. We also integrated Silero VAD \cite{silerovad} into the pipeline, which, with 400ms padding, minimized truncated transcriptions (as presented in Figure \ref{fig:mosel_transcription}) and reduced hallucinations by focusing inference on detected speech regions. Additionally, we modified the FasterWhisper source code to extract language IDs for each segment, enhancing our filtration pipeline.

\subsubsection{LID Verification}
We also noticed that eliminating data points where Whisper's predicted LID does not align with the target language significantly enhances the performance of the speech recognition model. 
We filtered out samples with multiple languages, common in the Voxpopuli dataset due to interpreter voices. For Granary's Voxpopuli set, we further refined filtering by excluding samples with low confidence Language ID predictions  ($\texttt{lid\_prob} < 0.8$).

\subsubsection{Robust Data Filtration}

Significant portion of filtration occurs at this stage of our pipeline, which involves three primary metrics for conducting the filtration process. First, we eliminate instances where any of the three hallucination flags are active, signaling the presence of \textit{i)} repeated \textit{n}-grams, \textit{ii)} long words, or \textit{iii)} frequently hallucinated phrases. The latter is particularly noteworthy; leveraging custom modifications to Whisper-large-V3 and V3-turbo, we compiled language-specific lists of commonly hallucinated phrases. These include both specific terms, such as \textit{"Sous-titrage Société Radio-Canada"} in French, and broadly used expressions like \textit{"Thank you very much"}. We use these lists to detect and filter hallucinated audio samples, and they will be made available as part of our open-source pipeline.

Character rate filtering is another crucial step. Using language- and corpus-specific heuristics, we eliminate speech-transcription pairs with anomalously low or high character rates, assuming such samples may contain non-speech segments, poor pseudo-labels, or unusual speech patterns.

Finally, we apply character set filtering by excluding any character deemed "invalid" for the Granary corpus. This comprehensive superset of over 300 characters and symbols ensures coverage of the alphabets across all 25 represented European languages.

\subsubsection{LLM-Powered P\&C Restoration}

The Granary corpus relies on pseudo-labeled data from Whisper, necessitating steps to enhance quality and reduce dependence on Whisper's performance. To address this, we applied punctuation and capitalization restoration using the large language model Qwen 2.5-7B-Instruct.

We crafted a language-specific prompt directing the model to assess and correct punctuation and capitalization, supplemented by multiple correction examples. To maintain output quality, we implemented character set filtering and Qwen hallucination filtering. A heuristic was used: if Qwen's output deviated from Whisper's transcriptions by more than a 5\% character error rate, the original pseudo-labels were retained. This margin allows Qwen to potentially correct typos or refine formulations. However, the quality of these modifications remains a subject for further testing.

\begin{figure*}[ht]
    \centering
    \caption{Total hours per language per task (ASR and AST) in Granary after final filtering (log scale). English has the most hours (275k for ASR), while Ukrainian has the least (932.67 for ASR, 608.80 for X$\rightarrow$En).}
    \includegraphics[width=1\linewidth]{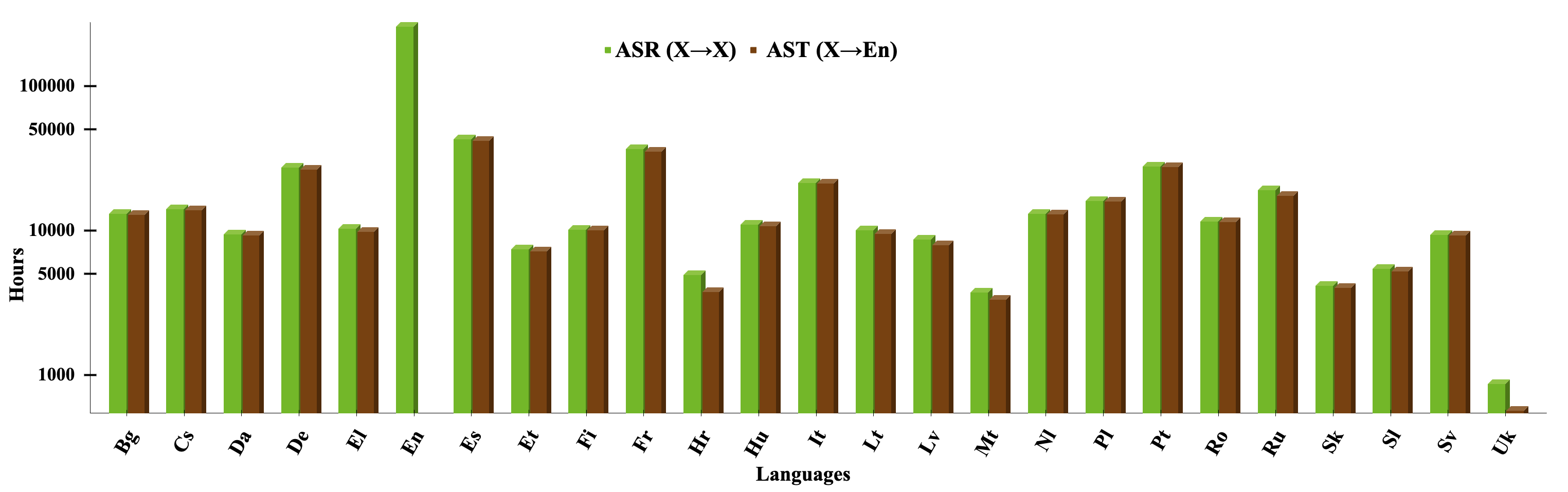}
    \label{fig:filtered_hrs_chart}
\end{figure*}

\subsection{AST Data Pipeline}

\subsubsection{Selection of Pseudo-Labeling Models}

We benchmarked several translation models to select the best model for X$\rightarrow$En AST pseudo labeling.
Our candidate models include LLMs such as Alma-13B-R \cite{xu2024a}, Qwen-2.5-7B \cite{qwen2}, EuroLLM-1.7B, and EuroLLM-9B \cite{DBLP:journals/corr/abs-2409-16235}, as well as encoder-decoder models such as 
Riva-Megatron Any2Any model\footnote{\url{https://catalog.ngc.nvidia.com/orgs/nvidia/teams/riva/models/riva_megatronnmt_any_any_1b}}. 
We excluded API-only models such as GPT-4o out of cost concerns, as well as TowerInstruct-13B and Aya-23 models because of non-commercial licenses.
Alma-13B-R and Qwen-2.5-7B were excluded during preliminary study, because the former under-performs significantly in speech domain (despite achieving impressive results on WMT test sets), and the latter suffers from hallucination issues during pseudo-labeling.
After a final comparison on the Flores dataset \cite{goyal-etal-2022-flores} covering all 24 translation directions of interest, we identified EuroLLM-9B as the best-performing model for AST data synthesis.

\subsubsection{LLM Inference}

We perform LLM inference on processed ASR data using the translation prompt from EuroLLM's model card.
For optimal speed, we use greedy inference with vLLM.
We also experimented with beam search, which provided a slight improvement for EuroLLM-1.7B but had diminishing returns for the 9B model, making the added computational cost unjustifiable.

\subsubsection{Filtration}

Prior work \cite{koehn-etal-2020-findings,peter-etal-2023-theres,finkelstein-etal-2024-introducing} has emphasized the importance of training data filtration for optimal translation performance.
Although our pseudo-labeled data is generated using an LLM optimized for translation, a small but persistent fraction of hallucinated examples remains.
This motivates us to develop an efficient and effective data filtration pipeline for our synthetic AST data.

Our data filtration pipeline is implemented in 
NeMo-Curator\footnote{\url{https://github.com/NVIDIA/NeMo-Curator/}},
a GPU-accelerated data curation toolkit.
Our filtration steps include a re-implementation of the length ratio filtering step from Moses
\footnote{\href{https://github.com/moses-smt/mosesdecoder/blob/master/scripts/training/clean-corpus-n.perl}{\url{https://github.com/moses-smt/.../clean-corpus-n.perl}}}
, character histogram filtering \cite{DBLP:journals/jmlr/FanBSMEGBCWCGBL21}, FastText language ID \cite{joulin2016bag}, as well as Quality Estimation filtration \cite{peter-etal-2023-theres}.
For the last step, we use \texttt{cometoid-wmt23} model \cite{gowda-etal-2023-cometoid} through PyMarian interface \cite{gowda-etal-2024-pymarian} as our quality estimation setup.
The resulting pipeline is very efficient and scalable.
For instance, when scaling over 8 nodes with 8 NVIDIA A100 GPUs each, filtering MOSEL dataset took only 47 minutes.

Overall, as presented in Table \ref{tab:corpora_hrs}, approximately 1 million hours of unlabeled data were processed to generate high-quality pseudo-labeled data, comprising approx 638,144 hours for ASR with a retention rate of 60.7\%, and 351,048 hours of X$\rightarrow$En AST pairs as part of Granary.

\section{Model Training and Evaluation}

In this section, we put the collected and processed data to use by training an ASR model. We focus on two languages: one high-resource language (English) and one low-resource language (Croatian). To evaluate the performance of our proposed pipeline, we use the filtered transcriptions provided by MOSEL \cite{mosel} as a baseline, which enables a direct comparison based on the same dataset. Our experiments utilize the FastConformer encoder \cite{rekesh2023fast} coupled with a hybrid RNNT-CTC decoder \cite{noroozi2023stateful}, employing the Large model configuration, which encompasses 120 million parameters.

The data utilized in this study is derived from VoxPopuli \cite{wang-etal-2021-voxpopuli}, with MOSEL \cite{mosel} providing pseudo-labeled transcriptions alongside metadata on hallucination features and language ID predictions generated by Whisper. We leveraged this information to create a filtered version of the MOSEL transcriptions for the VoxPopuli data. It is important to note that MOSEL published pseudo-labels only for a subset of Croatian VoxPopuli audio samples, comprising 2,800 transcribed hours out of a total of 8,000 available hours. To ensure a fair evaluation, we randomly sampled a comparable number of hours from Granary's VoxPopuli dataset in Croatian.

We evaluate our models on three test sets, both with and without punctuation and capitalization where applicable: VoxPopuli \cite{wang-etal-2021-voxpopuli} and FLEURS \cite{conneau2023fleurs} for English and Croatian. Since no validated test set is available for Croatian in Mozilla Common Voice (MCV), we conduct evaluations on MCV only for English. Additionally, we assess our models on the Hugging Face ASR leaderboard\cite{srivastav2023open} datasets for English.

All models are trained for 80,000 steps with a batch duration of approximately 10 hours per step, using 64 A100 80GB GPUs and a CosineAnnealing scheduler. The maximum learning rate is set to 1e-3 with a warmup of 15,000 steps and a constant weight decay of 1e-3. Model training is conducted using the NeMo framework\cite{nemo_toolkit} and Lhotse dataset modules \cite{zelasko2021lhotse}.

\begin{table}[ht]
    \centering
    \caption{WER of FastConformer-L on MOSEL and Granary English datasets [\%]}
    \resizebox{0.47\textwidth}{!}{%
    \begin{tabular}{l|c|c|c c|c c|c}
        \toprule
        \multirow{2}{*}{\textbf{Dataset}} & \multirow{2}{*}{\textbf{Hours}} & \multirow{2}{*}{\textbf{HF-Avg}} & \multicolumn{2}{c|}{\textbf{FLEURS}} & \multicolumn{2}{c|}{\textbf{MCV12}} & \textbf{VoxPopuli} \\
        \cmidrule(lr){4-5} \cmidrule(lr){6-7}
        & & & \textbf{PnC} & \textbf{noPnC} & \textbf{PnC} & \textbf{noPnC} & \textbf{noPnC} \\
        \toprule
        \midrule
        \textbf{MOSEL}   & 23,500  & 12.68  & 21.72  & 15.77  & 31.73  & 26.16  & 7.39 \\
        \textbf{Granary} & 14,000  & 12.57  & 19.63  & 13.93  & 31.32  & 26.40  & 7.25 \\
        \bottomrule
    \end{tabular}%
    }
    \label{tab:english_pnc}
\end{table}

As illustrated in Table \ref{tab:english_pnc}, we compare Granary with MOSEL, focusing on the performance of the ASR model trained on the filtered sets of both sources. Our goal is to create a high-quality pseudo-labeled dataset that retains approximately 50\% of the original 24,000 hours from the VoxPopuli English dataset. Remarkably, although the FastConformer model is trained on only 14,000 hours from the Granary dataset, it achieves results comparable to MOSEL's 23,500-hour filtered set. In fact, on most benchmarks, it slightly outperforms the larger dataset. Notably, we observe around a 10\% improvement on the highly reliable FLEURS test set, indicating that our more rigorous filtering process produces higher-quality training data compared to MOSEL \cite{mosel}.  A similar trend is observed on the Hugging Face ASR leaderboard as noted with HF-Avg (see Table \ref{tab:english_pnc}), further reinforcing our findings across different evaluation sets. For Croatian, we observe the same trend (see Table \ref{tab:croatian_pnc}), which indicate that our refined filtering methodology maximizes model performance within this particular setup, even with reduced data availability for both high- and low-resource languages.

\begin{table}[ht]
    \centering
    \caption{WER of FastConformer-L model on MOSEL and Granary Croatian datasets [\%]}
    \resizebox{0.3\textwidth}{!}{%
    \begin{tabular}{l|c|c|c|c}
        \toprule
        \multirow{2}{*}{\textbf{Dataset}} & \multirow{2}{*}{\textbf{Hours}} & \multicolumn{2}{c|}{\textbf{FLEURS}} & \textbf{VoxPopuli} \\
        \cmidrule(lr){3-4} 
        & & \textbf{PnC} & \textbf{noPnC} &  \textbf{noPnC}\\
        \toprule
        \midrule
        \textbf{MOSEL}   & 2,700  & 22.86  & 17.90  & 20.77 \\
        \textbf{Granary} & 2,100  & 21.75 & 17.14  & 20.38\\
        \bottomrule
    \end{tabular}%
    }
    \label{tab:croatian_pnc}
\end{table}

\section{Conclusion}

In conclusion, we present Granary, a comprehensive, open-source speech processing pipeline with transcriptions for speech recognition and translation across 25 European languages. Granary employs pseudo-labeling to enhance noisy public speech corpora, integrating open-source datasets and processes like audio segmentation, two-pass inference, language ID, robust data filtration, and LLM-based punctuation/capitalization restoration. Experiments on English and Croatian data show Granary's filtering improves model performance over existing datasets. Future work will focus on releasing multi-task, multilingual models trained on the complete Granary corpora.

\newpage

\bibliographystyle{IEEEtran}
\bibliography{mybib}

\end{document}